\documentclass[runningheads]{llncs}

\usepackage{eccv}

\usepackage{eccvabbrv}

\usepackage{graphicx}
\usepackage{booktabs}
\usepackage{amsmath}
\usepackage{wrapfig}

\usepackage[accsupp]{axessibility}  

\usepackage{hyperref}

\usepackage{orcidlink}
\usepackage{bbding}

\begin{document}

\title{Latent Visual Diffusion Reasoning \\ with Monte Carlo Tree Search} 

\titlerunning{LVDR}

\author{Xirui Teng\inst{1}\orcidlink{0000-1111-2222-3333} \and
Nan Xi\inst{2} \Envelope \orcidlink{0000-0002-7334-7772} \and
Junsong Yuan\inst{3}\orcidlink{0000-0002-7324-7034}} 

\authorrunning{X. Teng et al.}

\institute{Beijing Jiaotong University, Beijing, China \and
Virginia Commonwealth University, Richmond VA, USA \and 
University at Buffalo, Buffalo NY, USA \\
\Envelope \ \ \texttt{Corresponding author:} \email{xin@vcu.edu}
}

\maketitle

\begin{abstract}
  Analyzing fine-grained skill activities (e.g., sports, surgery) requires not only recognizing visual patterns but also performing step-by-step visual reasoning that leads to the final judgment. While recent advances in action quality assessment have achieved remarkable progress in evaluating performance, existing models remain black boxes, where they lack the ability to explicitly reveal the reasoning processes underlying their judgments. To address this limitation, we propose \textbf{Latent Visual Diffusion Reasoning (LVDR)}, a novel framework that integrates keypoint-guided Monte Carlo Tree Search (MCTS) to model and visualize the latent visual reasoning process. LVDR not only produces more accurate skill assessments but also uncovers the critical visual reasoning sequences that contribute to the final evaluation. Extensive experiments across four datasets spanning diverse sports and surgical domains demonstrate that LVDR achieves competitive quantitative performance while providing interpretable visual reasoning trajectories leading to the final predictions. Source codes and models can be found through the following link: \url{https://github.com/XiruiTeng/LVDR_Official.git}. 
\end{abstract}

\section{Introduction}

Assessing fine-grained skill activities such as athletic performances and surgical procedures poses a fundamental yet challenging problem in computer vision \cite{ashutosh2025expertaf, ashutoshlearning, xu2024vision, zhao2025pp, zhou2025phi, han2026caflow}. Unlike conventional action recognition \cite{xi2023open, xi2022forest, xi2023chain, wang2024interaction}, which focuses solely on identifying what action is being performed, skill assessment requires understanding how well the action is executed. This subtle distinction introduces a higher level of complexity, as it demands not only precise recognition of visual patterns but also the ability to perform \textbf{step-by-step visual reasoning} that links low-level motion dynamics to high-level evaluative outcomes. Capturing this causal progression from subtle temporal cues and biomechanical coordination to overall performance quality is critical for achieving meaningful and interpretable assessments. This capability is particularly important for high-stakes applications requiring transparency and trust, including athlete training, surgical education, and autonomous robotic skill acquisition.

\begin{figure}[t]
\centering
\includegraphics[width=0.93\textwidth]{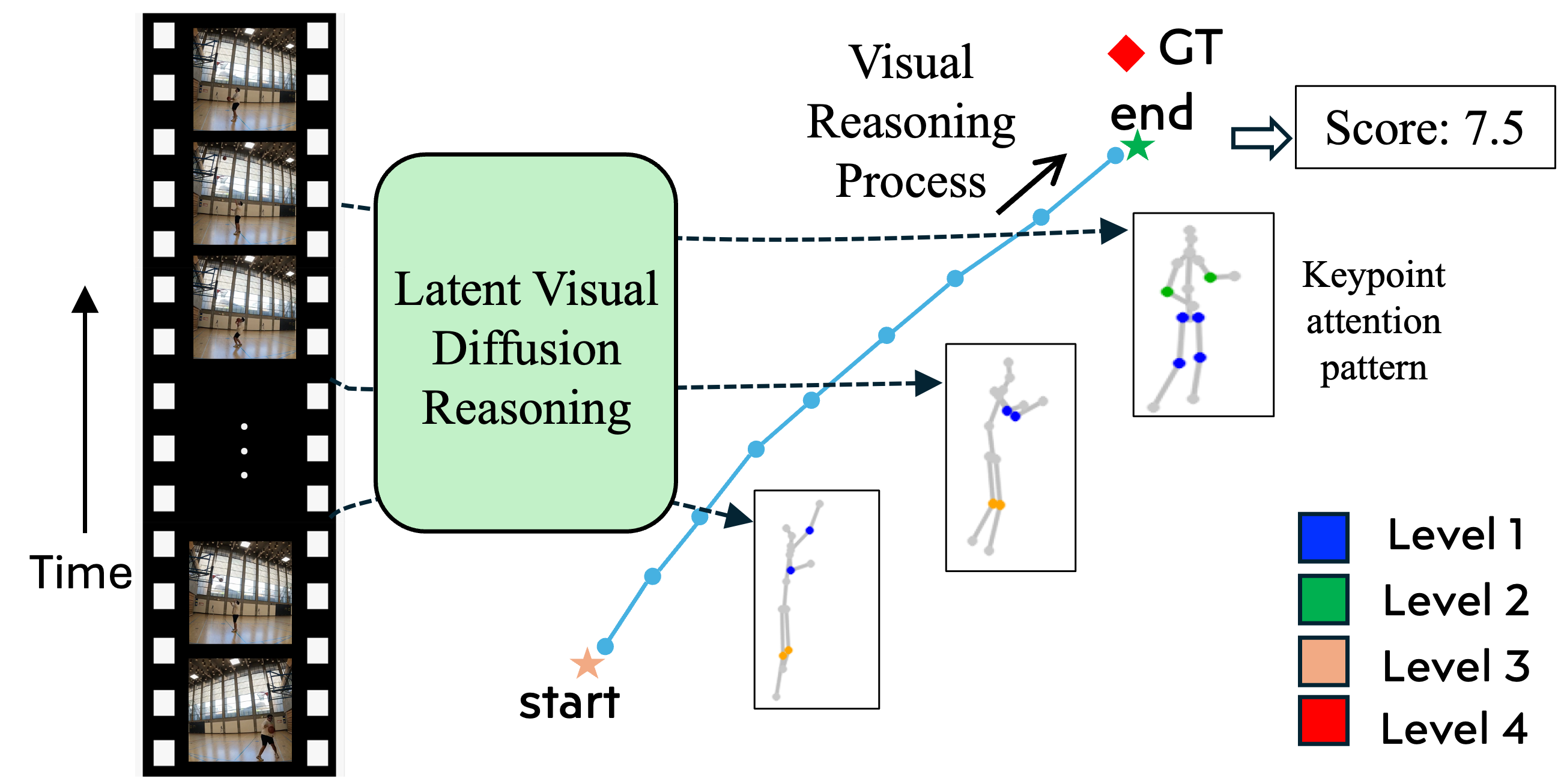} 
\caption{\textbf{Overview of Latent Visual Diffusion Reasoning.} We formulate visual reasoning as a diffusion process in the latent reasoning space. Our latent visual diffusion reasoning model is able to generate both  performance score for a given video and also the visual reasoning process with the evolving body keypoint attention pattern. Different colors on keypoints indicate different attention level at each time stamp.}
\label{fig:fig1}
\end{figure}

Recent advances in action quality assessment (AQA) \cite{MAGR, bai2022action, majeedi2024ricaˆ, yun2024semi} have demonstrated remarkable progress by leveraging deep neural networks to either regress continuous skill scores or predict relative performance rankings. While these models achieve strong predictive accuracy, they largely function as black boxes, which are capable of producing final assessments without exposing the intermediate reasoning steps that lead to those judgments. Consequently, their decision making processes remain opaque, limiting interpretability and hindering the ability of human experts to validate or derive actionable insights from model outputs. This lack of transparency poses a significant barrier to deployment in domains such as surgery and elite sports, where understanding \textit{why} a particular assessment is made is as important as the assessment itself. In such high-stakes settings, accountability, interpretability, and trust are indispensable for real-world adoption of AI-based assessment systems.

To address these limitations, we propose Latent Visual Diffusion Reasoning (LVDR), a novel framework that explicitly models the latent reasoning process underlying fine-grained skill assessment. As illustrated in Figure~\ref{fig:fig1}, LVDR not only predicts the overall performance score for a given action but also generates a visual reasoning trajectory that reveals how the model arrives at its judgment. Specifically, LVDR provides interpretable visualizations that capture the evolving keypoint attention patterns throughout the reasoning process. The framework integrates the representational power of diffusion-based generative modeling with the structured exploration capability of Monte Carlo Tree Search (MCTS).

We formulate video reasoning as a generative diffusion process within a latent space. LVDR treats reasoning as a progressive denoising process that evolves over temporal states. This latent space is structured by a target semantic distribution derived from ground-truth video semantics (e.g., labels or expert commentary). LVDR learns to generate coherent latent trajectories that represent a gradual refinement of semantic understanding. This formulation enables the model to capture the incremental and temporal nature of complex reasoning by explicitly modeling the path from a non-informative prior to a target semantic state. 

In reasoning space, the beginning of the video is treated as a noisy initialization of the reasoning process. As time progresses, the diffusion model gradually denoises the latent reasoning embeddings of the early video frames, transforming them into a target distribution that reflects coherent reasoning dynamics. Guided by keypoint-based MCTS, LVDR explores and identifies the most informative reasoning trajectories that contribute to the final evaluation. This design enables LVDR to make the model’s decision-making process both transparent and interpretable, bridging the gap between high predictive performance and human-understandable reasoning in fine-grained skill assessment.

To comprehensively evaluate the effectiveness of LVDR, we conduct experiments across four benchmark datasets spanning diverse skill domains, including multiple sports and surgical procedures. Experimental results demonstrate that LVDR achieves competitive or superior quantitative performance compared to state-of-the-art approaches, while uniquely offering interpretable visual reasoning trajectories that uncover key spatiotemporal patterns underlying skill execution. These results highlight LVDR’s ability not only to assess performance accurately but also to reveal the visual evidence and reasoning processes behind its judgments, paving the way toward more transparent and trustworthy skill assessment models.

In summary, our contributions include:
\begin{itemize}
    \item [$\cdot$] We introduce Latent Visual Diffusion Reasoning (LVDR), a framework that models step-by-step visual reasoning as a progressive diffusion process. The diffusion process is performed within a latent reasoning space constructed from the essential semantics of the full video, enabling coherent and interpretable reasoning for fine-grained skill assessment. 
    \item [$\cdot$] We propose a keypoint-guided Monte Carlo Tree Search mechanism to uncover and visualize latent visual reasoning sequences.
    \item [$\cdot$] We demonstrate that LVDR achieves strong performance across multiple sports and surgical datasets while offering unprecedented transparency in its decision-making process.
\end{itemize}

\section{Related Work}
\subsection{Action Quality Assessment} 
Action Quality Assessment(AQA) aims to evaluate how well an action is performed. This task is challenging for fine-grained skill activities, which require models to move beyond predicting scores and perform step-by-step visual reasoning. While this field has seen progress in predicting scores accurately by treating the task as a regression or ranking problem, most existing methods are mainly based on end-to-end deep learning methods and rarely reveal the processes which lead to their judgments. Recent efforts have attempted to address this problem. NS-AQA\cite{NS-AQA} utilizes neural networks to extract visual symbols from video data and rule-based models to analyze these symbols to generate detailed interpretations. IRIS\cite{Iris} uses scoring rubrics to segment action sequences and compute technical scores and this can give the explanations of the evaluation process. While these works make the reasoning process transparent, they are built on human-defined rules and fail to learn from data. Interpretability-AQA\cite{Interpretability-AQA} proposes a new attention loss function to avoid temporal skipping and a weight-score regression module to explain each video segment's score by giving different weights. While this method can explain how much a video segment contributes to the final score, it still fails to model the sequential, step-by-step reasoning process that a human expert might follow. To address this gap, we introduce a new framework that allows us to model and visualize the latent visual reasoning process, uncovering the critical reasoning sequences that lead to a final judgment while preserving quantitative accuracy. \\

\subsection{Monte Carlo Tree Search}
Monte Carlo Tree Search (MCTS) is a planning algorithm that strategically balances exploration and exploitation by integrating tree search with stochastic simulations \cite{coulom2006efficient}. The algorithm iteratively refines a search tree through four key stages. First, the selection phase traverses the tree from the root node, guided by a policy like Upper Confidence Bounds for Trees (UCT) \cite{kocsis2006bandit}, to select a child node that either maximizes potential score or explores less visited states. Upon reaching a leaf node or an expandable internal node, the expansion phase adds one or more child nodes, representing previously unconsidered actions or plan extensions. Subsequently, the simulation (or rollout) phase estimates the value of the newly added node by performing random or policy-based sampling of actions until a terminal state or a predefined depth is reached. Finally, the backpropagation phase updates the value estimates of the nodes along the path from the newly simulated node back to the root, incorporating the simulation outcome. Through repeated iterations of these four phases, MCTS adaptively focuses its search on promising regions of the state space while maintaining exploration of less certain areas.

\section{Method}
\label{sec:method}

\begin{figure*}[!h]
\centering
\includegraphics[width=0.98\textwidth]{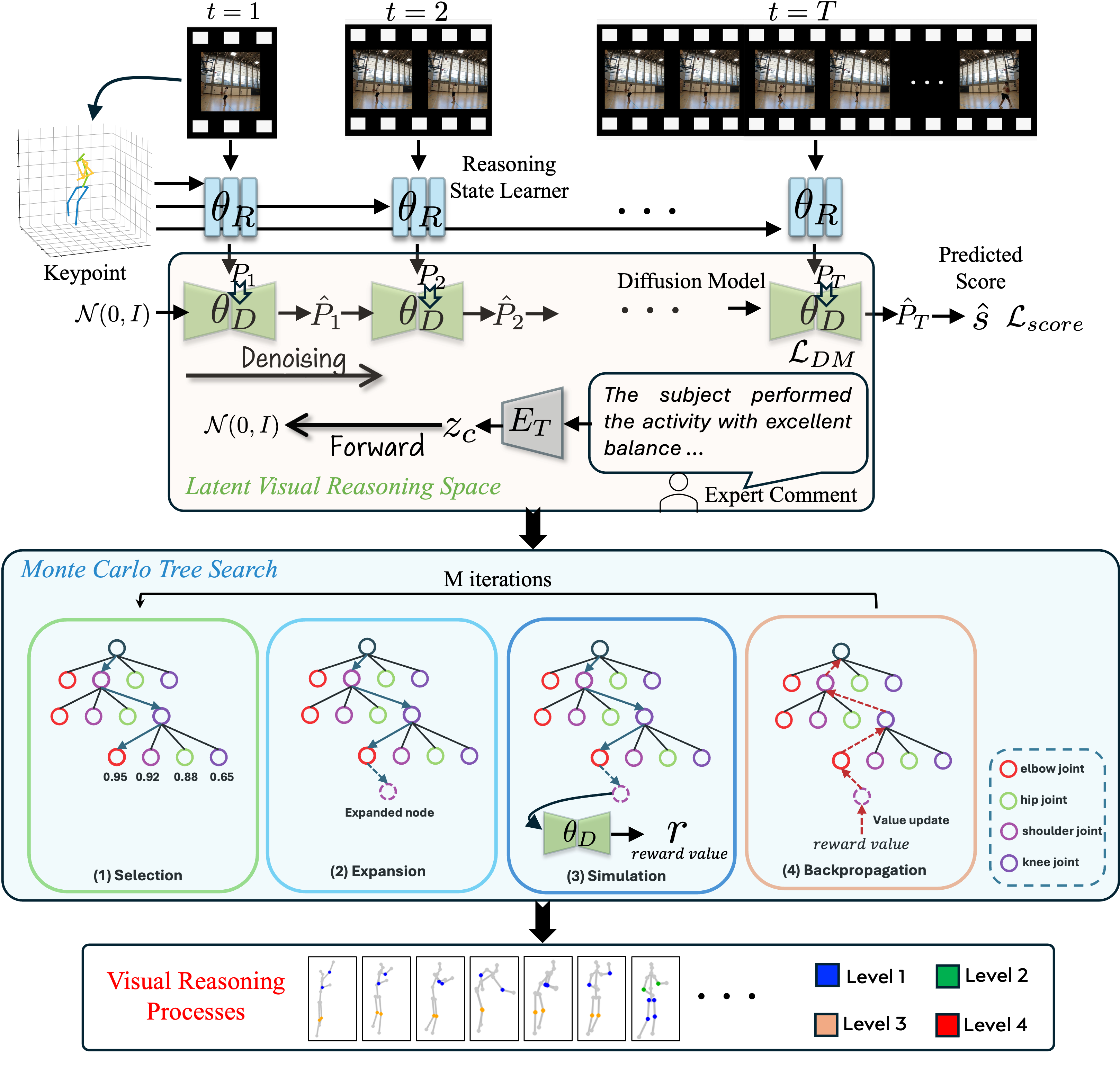} 
\caption{\textbf{Overall architecture of Latent Visual Diffusion Reasoning framework.} LVDR framework consists of two major components: (1) Latent Visual reasoning space, where we learn visual reasoning embeddings of the given video; (2) Monte Carlo Tree Search to extract explicit visual groundings that correspond to the most influential reasoning steps leading to the final prediction. Keypoint colors represent different levels of attention in the reasoning process. More detailed illustration can be found in Section~\ref{sec:method}.}
\label{fig:archtecture}
\end{figure*}

For a given video $V=\{I_1, \cdots, I_T \}$ that contains T video frames $\{I_i \}_{i=1}^T$, our LVDR framework $\mathcal{F}_{\theta}$ aims to generate: (1) predicted score $\hat{s}$; (2) visual reasoning process $\mathcal{H}=\{H_1, \cdots, H_T \}$, where $\{ H_i \}_{i=1}^T$ indicate keypoint attention patterns at each timestamp. LVDR includes two major components: (1) diffusion module to learn latent visual reasoning embeddings by modeling the denoising process of temporal reasoning states; (2) Monte-Carlo Tree Search (MCTS) that performs structured exploration in the latent reasoning space to discover informative visual reasoning trajectories.  

\subsection{Latent Visual Diffusion Reasoning}
Visual reasoning aims to provide a detailed illustration of the decision-making process that leads to the final prediction. However, in fine-grained video-based skill assessment, constructing such reasoning is highly challenging because frame-level annotations describing the reasoning process are nearly impossible to obtain. To address this limitation, we formulate visual reasoning as a diffusion process in a latent reasoning space, where the objective is to learn the target distribution of the reasoning process for the entire video. 

Our diffusion model is constructed based on Denoising Diffusion Implicit Models (DDIM) \cite{song2020denoising}. As shown in Figure~\ref{fig:archtecture}, we define a forward diffusion process $q$ that gradually adds Gaussian noise to clean text embedding $z_c$ over $T$ discrete time steps, where $z_c$ is derived from expert commentary using a frozen text encoder $E_T$. During training, we uniformly sample a random step $t \sim \mathcal{U}(\{1, \dots, T\})$. The noised latent $z_t$ at step $t$ can be sampled directly by using the reparameterization:
\begin{equation}
    z_t = \sqrt{\bar{\alpha}_t} z_c + \sqrt{1 - \bar{\alpha}_t} \epsilon, 
\end{equation}
where $\epsilon$ is a noise sampled from Gaussian distribution, $\bar{\alpha_t}$ is a noise schedule. We define $\bar{\alpha_t}$ with a cosine function:
\begin{equation}
    \bar{\alpha_t} = \cos\left(\frac{(t + 0.5)/T + s}{1+s}\cdot\frac{\pi}{2}\right)^2, 
\end{equation}
where $s$ is a small offset to prevent $\bar{\alpha_t}$ from approaching zero. 

Specifically, each video frame $I_i$ is associated with a corresponding latent visual reasoning embedding $\hat{P}_i$. At the beginning of the video, the reasoning embedding $P_1$ is inherently noisy due to the limited contextual information available from the first frame. As the video progresses, the diffusion model iteratively denoises these latent embeddings over time, producing progressively refined reasoning representations that converge toward the target distribution $P_T$. This formulation enables LVDR to model the evolution of reasoning as a continuous denoising trajectory, effectively capturing how the model’s internal reasoning process becomes more coherent and goal-directed as it observes more temporal context. 

We first extract 3D keypoints $\{K_i \}$ for each video frame $I_i$. Then we take $\{K_i \}$ and $I_i$ as inputs to the Reasoning State Learner $\theta_R$ to generate the initial visual reasoning embedding $P_i=\theta_R (I_I, K_i) \in \mathbb{R}^d$, where $d$ is the dimension of visual reasoning space. Then we employ a diffusion model $\theta_D$ to learn the actual visual reasoning embedding $\hat{P}_i=\theta_D (\hat{P}_{i-1}, P_i)$ based on its previous visual reasoning embedding $\hat{P}_{i-1} \in \mathbb{R}^d$ and conditioned on its own initial embedding $P_i$. Diffusion model $\theta_D$ is implemented with a diffusion transformer architecture and all $\theta_D$ at every timestamp share the same weights. The learning objective of diffusion model $\theta_D$ is:
\begin{equation}
    \mathcal{L}_{DM} = \mathbb{E}_{\epsilon \sim \mathcal{N}(0, 1), t, \hat{P}_t} \left[ || \epsilon-\epsilon_{\theta} (P_t, t, \hat{P}_t) ||^2 \right ], 
\end{equation}
where $\epsilon$ denotes noise and $\epsilon_{\theta}$ indicates denoising autoencoder. 


With the learned visual reasoning embedding $\hat{P}_t$, we are able to obtain the predicted score $\hat{s}=\mathrm{FC}(\hat{P}_t) \in \mathbb{R}$, where $\mathrm{FC}$ represents a fully connected layer. The loss function for regressing the predicted score is:
\begin{equation}
    \mathcal{L}_{score} = ||s-\hat{s}||^2, 
\end{equation}
where $s$ denotes the ground-truth score of the given video.
Thus, the total training objective is:
\begin{equation}
    \mathcal{L} = \mathcal{L}_{DM} + \lambda\mathcal{L}_{score}, 
    \label{eq:loss}
\end{equation}
where $\lambda$ indicates the hyperparameter for weight control and we take it as $1$ for implementation. 

\subsection{Monte-Carlo Tree Search (MCTS)}
With the proposed latent visual diffusion reasoning module, we obtain a sequence of latent reasoning embeddings that capture the model’s step-by-step visual reasoning process. The remaining challenge is to transform this implicit reasoning trajectory into an explicitly interpretable form. To address this, we incorporate Monte Carlo Tree Search (MCTS), a principled strategy for balancing exploration and exploitation in sequential decision-making. By integrating MCTS into our framework, we identify and extract explicit visual groundings that correspond to the most influential reasoning steps leading to the final prediction. This design not only enhances interpretability but also provides a transparent and structured view of how the model derives its decisions. 

Human referees rely on structured, sequential visual-processing strategies to accurately assess a performer’s actions, systematically attending to different regions of interest over time. For example, a referee may first focus on upper-body kinematics before shifting attention to lower-limb movements, forming a final judgment based on this ordered sequence of visual fixations. Inspired by this temporal and selective nature of human visual perception, we model these visual-processing strategies as combinatorial search spaces. Specifically, we formulate the referee’s visual process as a structured search problem and employ a keypoint-guided MCTS to capture and predict their dynamic attention patterns. For each video frame, we construct a MCTS, as illustrated in Figure~\ref{fig:archtecture}, to identify the most critical keypoints that contribute to the visual reasoning process.

In the constructed MCTS, each node represents a state (represented with circles of different colors in Figure~\ref{fig:archtecture}), each edge represents a transition from one state to another. To start with, we first define the tree nodes based on the particular problem we are solving. In sports scenarios, we define tree nodes based on the four major joints that control human movement: \textit{elbow}, \textit{shoulder}, \textit{hip}, and \textit{knee}. Therefore, every tree node in sports scenarios consists of four possible actions. In surgical scenarios, since surgical actions are performed via surgical instruments, we define tree nodes based on the three major parts of instruments: \textit{tip}, \textit{body} and \textit{tail}. Therefore, every tree node in surgical scenarios consists of three possible actions. We use sports scenarios as running examples and include illustrations on surgical scenarios in supplement materials. 

For the MCTS of each video frame, we implement the keypoint-guided tree with a finite depth $K$. Each leaf node involves $K-1$ steps and corresponds to a unique trajectory through the assessment process, representing a sequence of consecutive. To represent each node, we utilize the three key physical information of each joint: \textit{angle}, \textit{speed}, and \textit{position}. Formally, each node is represented in the form of $(a, v, p)$ where $a \in [0,1]$ indicates angle, $v \in \mathbb{R}$ denotes the joint speed and $p=(x, y, z)$ indicates the position of the joint, where $x\in[0,1]$, $y\in[0,1]$ and $z\in[0,1]$ are normalized 3D coordinates. This representation models the hierarchical relationships between major joints and their associated kinematic properties, providing a structured representation of the domain knowledge relevant to skilled human activities. 

The decision tree of judgment is constructed iteratively when MCTS is executed over $M$ iterations at each judgment step. Each iteration includes four phases as shown in Figure~\ref{fig:archtecture}: 

\noindent \textbf{1. Selection:} Starting from the current root node, we traverse the tree by repeatedly selecting a child node according to the standard Upper Confidence Bound for Trees (UCT) criterion, which balances exploration and exploitation. This process continues until reaching a leaf node or a partially expanded node, ensuring that the search prioritizes both promising and underexplored branches.

\noindent \textbf{2. Expansion:} Upon reaching such a node, the expansion step generates new child nodes by exploring previously unvisited paths that partition the remaining possibility space. This operation incrementally refines and narrows the search space, enabling the tree to progressively represent a richer set of candidate reasoning trajectories. 

\noindent \textbf{3. Simulation:} A rollout policy is employed to estimate the expected reward of the selected node by leveraging the learned diffusion model $\theta_D$. Beginning from this node, the policy simulates a sequence of interactions until either a terminal state is reached or a predefined rollout depth is satisfied. This simulation provides an approximate evaluation of the node’s downstream potential within the reasoning space.

\noindent \textbf{4. Backpropagation:} Once the simulation concludes at a leaf node u, the resulting expected reward is propagated backward through the tree. All ancestor nodes along the path from u to the initially selected node are updated accordingly, ensuring that both favorable and unfavorable simulated outcomes influence future selection decisions.

After $M$ iteration, the selected path $v^*$ has the highest expected reward:
\begin{equation}
    v^* = \arg max _{v'} R_e(v'), 
\end{equation}
where $R_e$ indicates the expected reward of the given path. 

During inference, we first apply the trained LVDR model to the test videos to obtain the corresponding latent visual reasoning embeddings. We then perform MCTS on these embeddings to derive explicit keypoint-based visual groundings that reveal the model’s step-by-step reasoning process.

\section{Experiments}
\vspace{-1mm}

\subsection{Datasets}
We evaluate our Latent Diffusion Reasoning model on four datasets: (1) \textbf{EgoExo4D}\cite{EgoExo4d} dataset comprises multi-view videos of daily and sports activities; for sports skill assessment, we use three action categories: basketball, soccer and climbing; (2) \textbf{JIGSAWS}\cite{JIGSAWS} dataset comprises videos of robot-assisted surgical tasks; used for surgical skills assessment; (3) \textbf{FitnessAQA}\cite{fitnessaqa} dataset comprises videos with expert annotations to evaluate the performance of athletes; used for fitness skills assessment; (4) \textbf{Cataract-101}\cite{schoeffmann2018cataract} dataset comprises videos of cataract surgeries; used for cataract surgical skill assessment. Detailed introduction of datasets can be found in supplement materials.

\begin{table}[t]
    \centering
    \resizebox{0.85\linewidth}{!}{
    \begin{tabular}{l|cc|cccc}
    \toprule
    \textbf{Models} & \multicolumn{2}{c|}{\textbf{JIGSAWS}} & \multicolumn{4}{c}{\textbf{FitnessAQA}} \\
    \cmidrule(lr){2-3}\cmidrule(lr){4-7} & $\rho \uparrow$ & $R-\ell_2(\times 100)\,\downarrow$ & F1 (OK) & F1 (OE) & F1 (SF) & F1 (SI) \\
    \midrule
    MAGR\cite{MAGR} & 0.48 & 10.60 & 0.564 & 0.357 & 0.796 & 0.157 \\
    MAGR++\cite{MAGRPP} & 0.55 & 10.20 & 0.485 & 0.022 & \textbf{0.818} & 0.000\\
    LLaVA-Video\cite{LLAVA-Video} & 0.41 & 40.61 & 0.392 & 0.533 & 0.795 & 0.257 \\
    Gemini 2.5\cite{gemini25} & 0.10 & 33.13 & 0.105 & 0.128 & 0.513 & 0.170 \\
    GPT-4o\cite{gpt4o} & 0.09 & 71.61 & 0.000 & 0.046 & 0.000 & 0.000 \\
    Qwen-2.5-VL-7B\cite{Qwen} & 0.03 & 67.48 & 0.404 & 0.082 & 0.000 & 0.000 \\
    \midrule
    Ours($\Delta$ Diffusion) & 0.65 & 10.73 & \textbf{0.768} & 0.497 & 0.725 & \textbf{0.920} \\
    \textbf{Ours} & \textbf{0.69} & \textbf{9.93} & 0.747 & \textbf{0.818} & 0.813 & \textbf{0.920} \\
    \midrule
    \end{tabular}}
    \vspace{2mm}
    \caption{\textbf{Action score prediction results in JIGSAWS and FitnessAQA dataset.} OK: OHP Knee; OE: OHP Elbow; SF: Squat Forward; SI: Squat Inward.
    }
    \label{tab:JIGSAW_FitnessAQA_sota}
\end{table}

\subsection{Evaluation Metrics}
\begin{wraptable}{l}{0.5\textwidth} 
    \centering
    \resizebox{0.48\textwidth}{!}{
        \begin{tabular}{l|c|c}
        \toprule
        \textbf{Models} & $\rho \uparrow$ & $R-\ell_2(\times 100)\,\downarrow$ \\
        \midrule
        FineParser\cite{FineParser} & 0.41 & 29.36 \\
        MAGR\cite{MAGR} & 0.73 & 10.14\\
        MAGR++\cite{MAGRPP} & 0.71 & 4.17 \\
        Qwen-2.5-VL-7B\cite{Qwen} & 0.46 & 44.26 \\
        LLaVA-Video\cite{LLAVA-Video} & 0.21 & 16.93 \\
        GPT-4o\cite{gpt4o} & 0.54 & 11.87 \\
        Gemini 2.5\cite{gemini25} & 0.61 & 5.57 \\
        \midrule 
        Ours($\Delta$ Diffusion) & 0.83 & 2.60 \\
        \textbf{Ours} & \textbf{0.88} & \textbf{1.69} \\
        \midrule
        \end{tabular}
    }
\vspace{-1mm}
    \caption{\textbf{Action score prediction results in EgoExo4D dataset.}}
    \label{tab:ADR_Dataset_sota}
\end{wraptable}


\begin{figure*}[t]
\centering
\includegraphics[width=\textwidth]{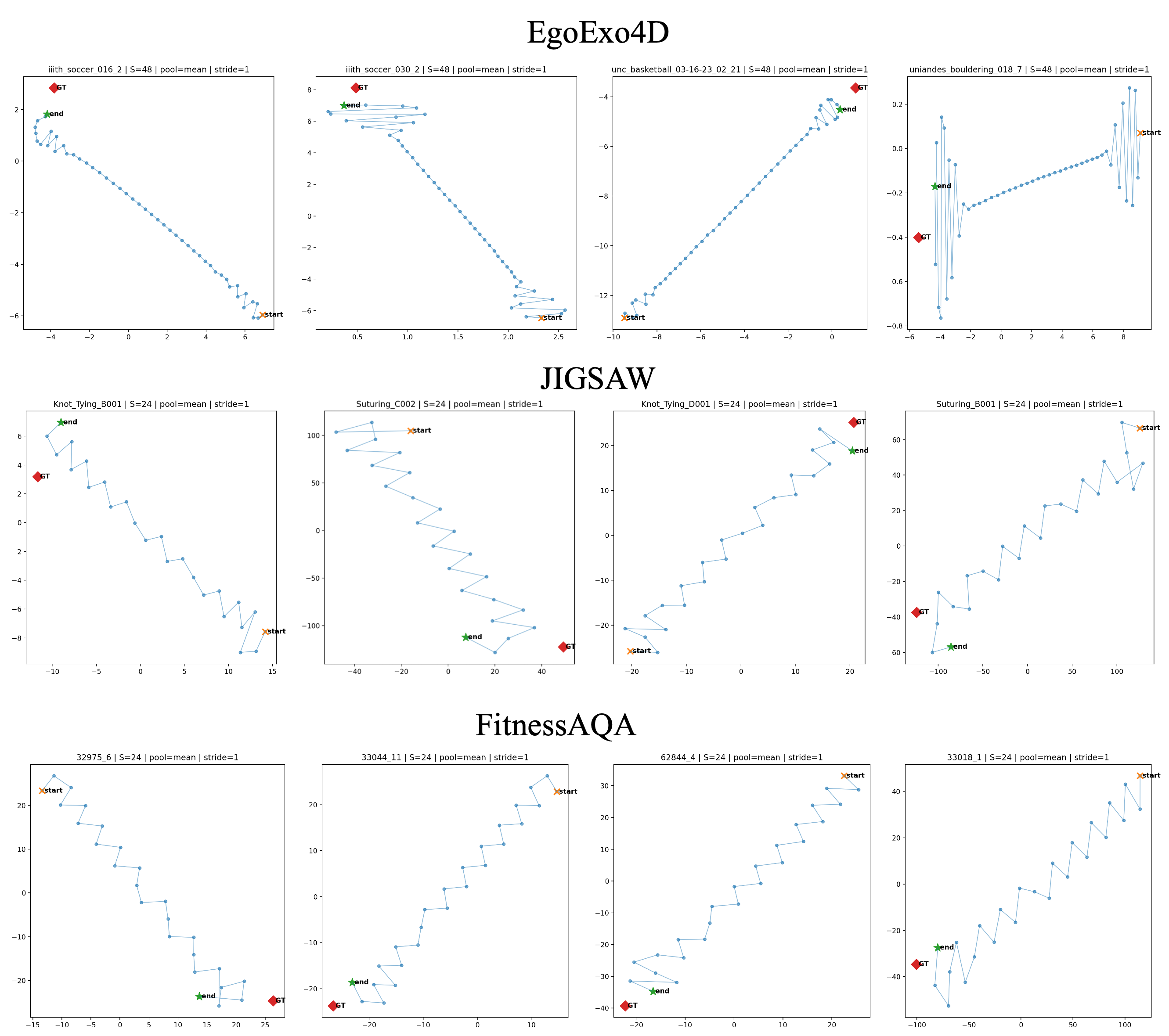} 
\vspace{-2mm}
\caption{\textbf{Visual reasoning trajectory visualization results.} We present visual reasoning trajectories on 2D space with three datasets. Trajectories show the progressive reasoning process from start point to end point.}
\label{fig:trajectory}
\end{figure*}
We employ two metrics to qualify the performance of models: Spearman’s Rank Correlation ($\rho$) \cite{xu2022finediving} and relative $l_2$ distance $R-\ell_2(\times 100)$ \cite{xu2022finediving}. For the FitnessAQA dataset, we adopt the F1-score following the prior work, while for the Cataract-101 dataset, model performance is assessed using accuracy.

\begin{figure}[!t]
\centering
\includegraphics[width=0.99\textwidth]{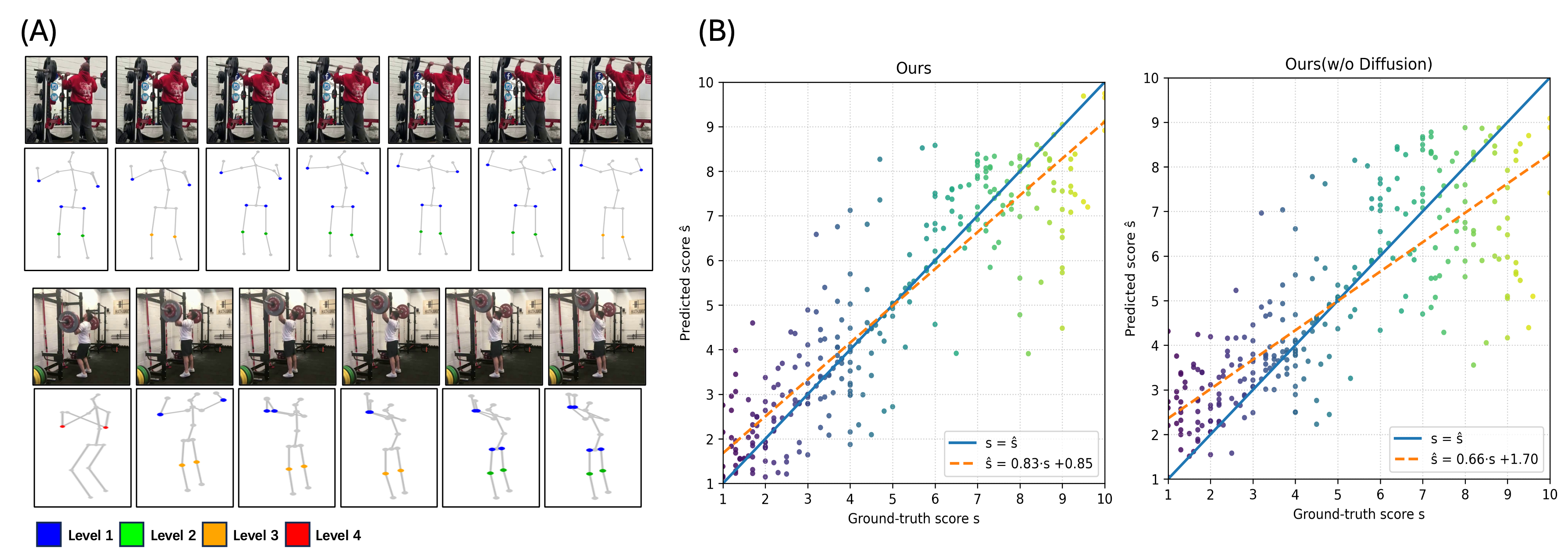} 
\caption{\textbf{(A) Keypoint visualization results of FitnessAQA dataset.} Keypoint colors represent different levels of attention in the reasoning process. \textbf{(B) Scatter plots on EgoExo4D dataset.} We compare the scatter plots for prediction score performance on EgoExo4D dataset with and without diffusion module.}
\label{fig:keypoint_fitnesaqa&scatter_plot}
\end{figure}

\subsection{Implementation Details}
\textbf{Training Details}. We utilize visual encoder and text encoder from InternVideo2.5\cite{InternVideo} to extract visual and text features. Our diffusion model is a Transformer-based architecture with 12 layers and 16 attention heads. The dimension of the hidden state is 512. We use Adam optimizer with a learning rate of $1e-4$ and a batch size of 8 to train our diffusion model and score-prediction module. \\

\noindent \textbf{MCTS Inference Details.} For the MCTS inference, the depth $K$ is set to 4 in the sports scenarios and 3 in the surgical scenarios. We use the Upper Confidence Bound for Trees(UCT) algorithm for node selection. The reward $r$ is calculated as the cosine similarity between the embedding of the path's keypoints and the pre-computed target embedding for that step. 
All experiments are conducted on a single NVIDIA L40S GPU. 

\subsection{Quantitative Results}


\begin{table}[t]
\centering

\resizebox{0.9\linewidth}{!}{
    \begin{tabular}{l|cc|cc|cc}
    \toprule
    \textbf{Models} & \multicolumn{2}{c|}{\textbf{Soccer}} & \multicolumn{2}{c|}{\textbf{Basketball}} & \multicolumn{2}{c}{\textbf{Climbing}}\\
    \cmidrule(lr){2-3}\cmidrule(lr){4-5}\cmidrule(lr){6-7} & $\rho \uparrow$ & $R-\ell_2(\times 100)\, \downarrow$ & $\rho \uparrow$ & $R-\ell_2(\times 100)\, \downarrow$ & $\rho \uparrow$ & $R-\ell_2(\times 100)\, \downarrow$\\
    \midrule
    MAGR\cite{MAGR} & 0.52 & 8.43 & 0.72 & 7.90 & 0.74 & 13.32\\
    MAGR++\cite{MAGRPP} & 0.59 & 4.44 & 0.58 & 6.54 & 0.67 & 7.45\\
    FineParser\cite{FineParser} & 0.71 & 45.52 & 0.12 & 24.43 & 0.72 & 13.18\\
    LLaVA-Video\cite{LLAVA-Video} & 0.07 & 29.64 & 0.31 & 13.58 & 0.48 & 20.23 \\ 
    GPT-4o\cite{gpt4o} & 0.18 & 38.64 & 0.51 & 10.17 & 0.75 & 16.7 \\
    Gemini 2.5\cite{gemini25} & 0.64 & 14.21 & 0.63 & 7.01 & 0.70 & 4.34 \\ 
    Qwen-2.5-VL-7B\cite{Qwen} & 0.70 & 69.49 & 0.48 & 11.37 & 0.43 & 59.67 \\
    \midrule
    Ours($\Delta$ Diffusion) & \textbf{0.97} & \textbf{0.67} & 0.78 & 4.13 & 0.83 & 3.18 \\
    \textbf{Ours} & 0.94 & 0.81 & \textbf{0.88} & \textbf{2.08} & \textbf{0.87} & \textbf{2.16} \\
    \midrule
    \end{tabular}
}
\vspace{2mm}
\caption{\textbf{Quantitative results of score prediction performance.} Experiments on the EgoExo4D dataset for the three sports categories (Soccer, Basketball, Climbing).}
\label{tab:Split_Results_on_ADR_dataset}
\end{table}

\begin{table}[t]
\centering

\resizebox{0.9\linewidth}{!}{
    \begin{tabular}{l|cc|cc|cc}
    \toprule
    \textbf{Models} & \multicolumn{2}{c|}{\textbf{Knot-Tying}} & \multicolumn{2}{c|}{\textbf{Needle-Passing}} & \multicolumn{2}{c}{\textbf{Suturing}}\\
    \cmidrule(lr){2-3} \cmidrule(lr){4-5} \cmidrule(l){6-7} & 
    $\rho \uparrow$ & $R-\ell_2(\times 100)\, \downarrow$ & $\rho \uparrow$ & $R-\ell_2(\times 100)\, \downarrow$ & $\rho \uparrow$ & $R-\ell_2(\times 100)\, \downarrow$ \\
    \midrule 
    MAGR \cite{MAGR} & 0.45 & 8.84 & 0.56 & 13.37 & 0.42 & 9.60 \\
    MAGR++ \cite{MAGRPP} & 0.50 & 9.86 & \textbf{0.60} & \textbf{11.10} & 0.55 & 9.64 \\
    LLaVA-Video \cite{LLAVA-Video} & 0.45 & 31.74 & 0.17 & 48.69 & 0.61 & 41.41 \\
    Gemini 2.5 \cite{gemini25} & 0.05 & 29.07 & 0.14 & 50.59 & 0.11 & 19.73 \\
    GPT-4o \cite{gpt4o} & 0.27 & 72.09 & 0.08 & 119.51 & 0.07 & 23.23 \\ 
    Qwen-2.5-VL-7B \cite{Qwen} & 0.47 & 72.81 & 0.26 & 108.88 & 0.29 & 20.73 \\
    \midrule
    Ours($\Delta$ Diffusion) & 0.77 & 5.89 & 0.53 & 20.15 & 0.64 & 6.15 \\
    \textbf{Ours} & \textbf{0.78} & \textbf{5.37} & \textbf{0.60} & 18.28 & \textbf{0.68} & \textbf{6.14} \\
    \midrule
    \end{tabular}   
}
\vspace{2mm}
\caption{\textbf{Quantitative results of score prediction performance on the JIGSAWS dataset.} Three surgical activities are included: Knot-Tying, Needle-Passing, Suturing.}
\label{tab:Split_Results_on_JIGSAWS_dataset}
\end{table}

\begin{figure*}[!t]
\centering
\includegraphics[width=0.98\textwidth]{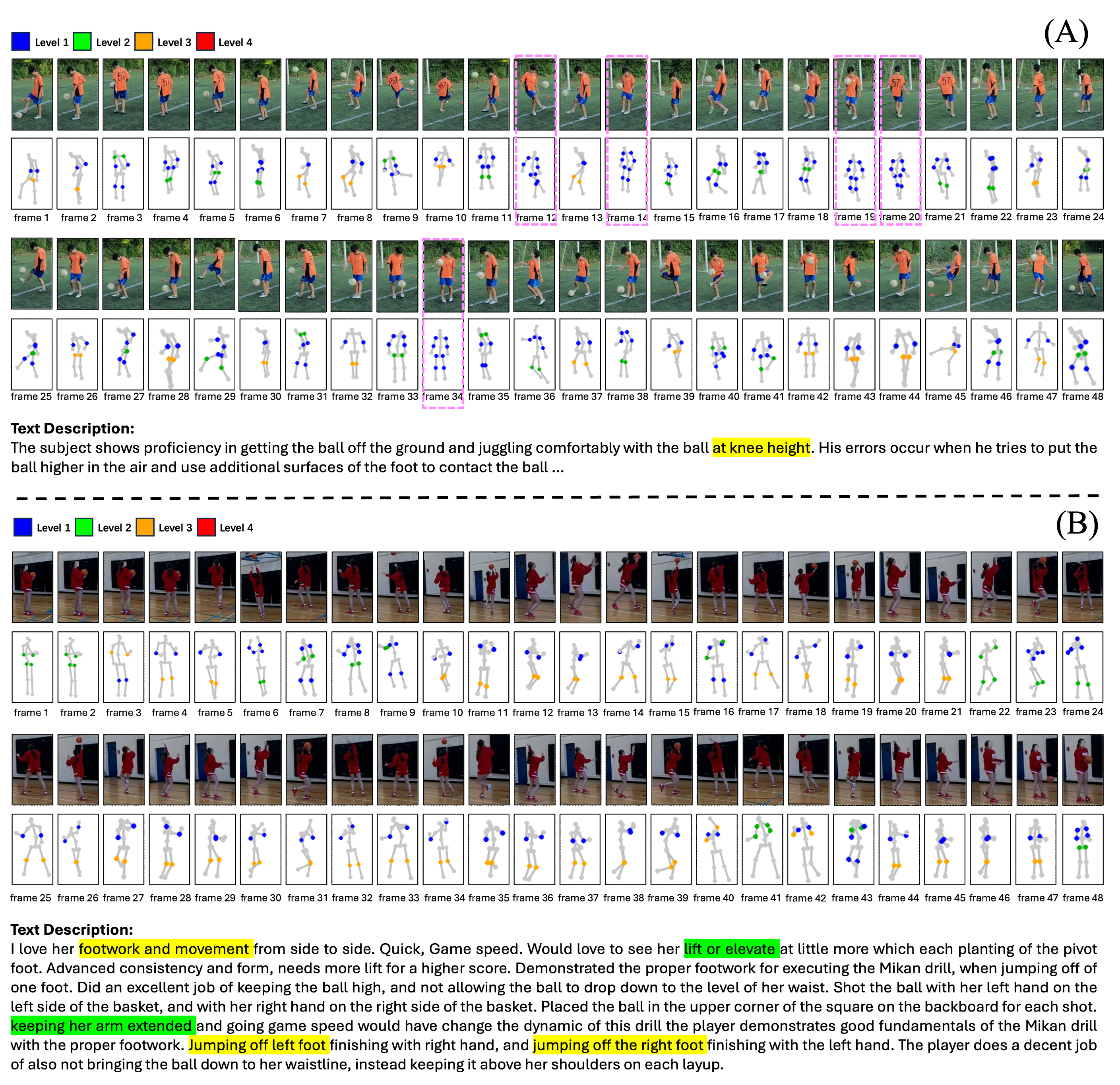} 
\caption{\textbf{Keypoint visualization results in soccer and basketball.} Keypoint colors represent different levels of attention in the visual reasoning process. Detailed illustration can be found in Section~\ref{sec:qualitative}.}
\label{fig:keypoint_all}
\end{figure*}

We evaluate the quantitative performance of our LVDR model on four datasets spanning both sports and surgical domains. As shown in Table~\ref{tab:ADR_Dataset_sota}, our model achieves state-of-the-art performance on the EgoExo4D dataset, outperforming both domain-specific approaches (e.g., MAGR~\cite{MAGR}) and large vision–language models (e.g., GPT-4o~\cite{gpt4o}) in predicting performance scores. Consistent improvements are also observed in Table~\ref{tab:JIGSAW_FitnessAQA_sota} and Table~\ref{tab:Catarat-101Results}, further demonstrating the robustness and generalizability of our LVDR framework across diverse video-based skill assessment benchmarks. In Table~\ref{tab:Split_Results_on_ADR_dataset} and Table~\ref{tab:Split_Results_on_JIGSAWS_dataset}, we also present score prediction performance of individual sports categories (soccer, basketball and climbing) and surgery actions (knot-tying, needle-passing and suturing). In those individual categories, our model also surpasses competing methods in most cases.

\subsection{Visual Reasoning Trajectory Evaluation}
\noindent \textbf{User Study.} We invited a doamin expert in soccer and basketbnall to evaluate the correctness of the keypoint reasoning processes generated by MCTS at each video frame. The expert assigned one of three labels: Correct (C); Wrong (W); Not Sure (N). We assess a total of 50 soccer and basketball videos and report percentage results of each category: $86\%$ C; $8\%$ W; $6\%$ N. These results indicate strong alignment between the generated keypoint reasoning processes and human expert judgment. \\

\noindent{\textbf{Keypoint Masking Experiments.}} To further verify that MCTS identifies critical motion cues for skill assessment, we conduct keypoint masking experiments from two complementary perspectives: \textbf{(1)} masking the keypoints selected by MCTS, and \textbf{(2)} masking the keypoints \textit{not} selected by MCTS.
\begin{wraptable}{l}{0.35\textwidth} 
    \centering
    \resizebox{0.3\textwidth}{!}{
        \begin{tabular}{l|c}
        \toprule
        \textbf{Models} & Acc$\,\uparrow$\\
        \midrule
        MAGR\cite{MAGR} & 0.62 \\
        MAGR++\cite{MAGRPP} & 0.62 \\
        Qwen-2.5-VL-7B\cite{Qwen} & 0.53 \\
        LLaVA-Video\cite{LLAVA-Video} & 0.48 \\ 
        \midrule
        Ours($\Delta$Diffusion) & 0.67 \\
        \textbf{Ours} & \textbf{0.71}\\
        \midrule
        \end{tabular}
    }
\vspace{-1mm}
    \caption{\textbf{Action score prediction results in Cataract-101 dataset}}
    \label{tab:Catarat-101Results}
\end{wraptable}
Table 6 in the main paper reports a subset of these results, while Table~\ref{tab:masking_MCTS_Ego} presents the complete evaluation from both perspectives. As shown in Table~\ref{tab:masking_MCTS_Ego}, masking the MCTS-selected keypoints ($\Delta$ MS) causes a substantial drop in model performance. In contrast, masking keypoints not selected by MCTS ($\Delta$ MUNS) results in only marginal or negligible performance degradation. These findings indicate that MCTS effectively identifies the keypoints that encode the most critical motion cues for skill assessment. \\

\begin{table}[!h]
    \centering
    \resizebox{0.97\linewidth}{!}{
    \begin{tabular}{l|cc||l|c||l|c }
        \toprule
        \textit{EgoExo4D} & $\rho \uparrow$ & $R-\ell_2(\times 100)\, \downarrow$ & \textit{Cataract-101}  & ACC $\uparrow$ & \textit{FitnessAQA} & F1 (OHP Elbow) $\uparrow$  \\
        \midrule
        \textbf{Ours} ($\Delta$ MS) & \textcolor{red}{\textbf{0.86}} & \textcolor{red}{\textbf{2.16}} & \textbf{Ours} ($\Delta$ MS) & \textcolor{red}{\textbf{0.53}} & \textbf{Ours} ($\Delta$ MS) & \textcolor{red}{\textbf{0.282}} \\
        \textbf{Ours} ($\Delta$ MUNS) & {0.88} & 1.78 & \textbf{Ours} ($\Delta$ MUNS) & {0.71} & \textbf{Ours} ($\Delta$ MUNS) & {0.818} \\
        \midrule
        \textbf{Ours} & {0.88} & {1.69} & \textbf{Ours} & {0.71} & \textbf{Ours} & {0.818} \\
        \bottomrule
    \end{tabular}
    }
    \vspace{2mm}
    \caption{Score prediction results for EgoExo4D, Cataract-101 and FitnessAQA datasets by (1) masking MCTS selected keypoints; (2) masking MCTS unselected keypoints. Results of masking MCTS selected keypoints are highlighted in \textcolor{red}{\textbf{red}}, indicating the difference with the original model performance. $\Delta$ MS: mask MCTS selected keypoints; $\Delta$ MUNS: mask MCTS unselected keypoints.}
    \label{tab:masking_MCTS_Ego}
\end{table}

\subsection{Inference Latency}
In Table~\ref{tab:inf_speed}, we report the inference speed per frame under different numbers of MCTS iterations. The results demonstrate that our model maintains \textbf{efficient inference} across these settings, demonstrating its suitability for real-time deployment with fast MCTS-based reasoning. 

\vspace{2mm}
\begin{table}[!h]
    \centering
    \resizebox{0.95\linewidth}{!}{
        \begin{tabular}{c|ccc} 
      \toprule
       \textbf{Ours} (w/o MCTS) & \textbf{Ours} (75 MCTS) & \textbf{Ours} (150 MCTS) & \textbf{Ours} (225 MTCS) \\
        \midrule
        0.006s & 0.048s & 0.094s & 0.129s \\
      \bottomrule 
    \end{tabular}
    }
\vspace{2mm}
    \caption{Inference speed per frame. The first column represents our method with only score prediction (no MCTS). Column 2-4 indicate our model with 75, 150 and 225 MCTS iterations, respectively.}
    \label{tab:inf_speed}
\end{table}


\subsection{Ablation Study}
Furthermore, we conduct an ablation study to assess the contribution of the diffusion module within our LVDR framework. As shown in Table~\ref{tab:ADR_Dataset_sota}, \ref{tab:JIGSAW_FitnessAQA_sota}, \ref{tab:Catarat-101Results}, \ref{tab:Split_Results_on_ADR_dataset} and \ref{tab:Split_Results_on_JIGSAWS_dataset}, removing the diffusion model leads to a consistent performance degradation across all datasets, confirming its essential role in modeling progressive visual reasoning. In addition, the scatter plots in Fig.~\ref{fig:keypoint_fitnesaqa&scatter_plot}-B further illustrate the performance gap between LVDR variants with and without the diffusion component, reinforcing the effectiveness of diffusion-based reasoning in our framework. 

In Table~\ref{tab:additional_ablations}, we report results of additional ablation studies including different loss function weights in Eq.~\ref{eq:loss} weights and noise scheduler in diffusion model. 
\vspace{2mm}
\begin{table}[!t]
    \centering
    \resizebox{0.9\linewidth}{!}{
    \begin{tabular}{l|cc || l|c}
        \toprule
        {EgoExo4D dataset} & $\rho \uparrow$ & $R-\ell_2(\times 100)\, \downarrow$ & {Catatact-101 dataset} & Acc $\uparrow$ \\
        \midrule
        $\lambda=0.5$ & 0.86 & 3.86 & $\lambda=0.5$ & 0.67 \\
        $\lambda=2$ & 0.86 & 2.07 & $\lambda=2$ & 0.67 \\
        $\lambda=1$ (\textbf{Ours}) & \textbf{0.88} & \textbf{1.69} & $\lambda=1$ (\textbf{Ours}) & \textbf{0.71} \\
        \midrule
        liner schedule & 0.85 & 2.25 & liner schedule & 0.67 \\
        cosine schedule (\textbf{Ours}) & \textbf{0.88} & \textbf{1.69} & cosine schedule & \textbf{0.71} \\
        \bottomrule
    \end{tabular}
    }
\vspace{2mm}
    \caption{Score prediction results with different loss weights and noise scheduler in diffusion model.}
    \label{tab:additional_ablations}
\end{table}

\subsection{Qualitative Results}
\label{sec:qualitative}
We visualize the latent visual reasoning trajectories in Figure~\ref{fig:trajectory}. Each trajectory depicts the temporal evolution of the model’s reasoning state from the initial step to the final prediction. At early steps, the reasoning state lies far from the ground-truth representation, reflecting higher uncertainty in the model’s understanding. As reasoning progresses, the trajectory moves steadily closer to the ground-truth point, indicating a gradual and consistent refinement of the inferred semantics. This visualization demonstrates that our LVDR framework successfully learns a coherent and interpretable reasoning path within the latent visual reasoning space. 

To assess the reliability of MCTS, we visualize the keypoint selection results in Figure~\ref{fig:keypoint_fitnesaqa&scatter_plot}-A and Fig.~\ref{fig:keypoint_all}. The color of each keypoint reflects how many times it is selected during the MCTS procedure: Level 1 indicates a single selection, whereas Level 4 indicates four selections. We further highlight in the text the essential semantic cues that correspond to these visualizations (e.g., the description ``footwork and movement'' in Figure~\ref{fig:keypoint_all} (B) aligns with the strong attention on the knee joints). In Figure~\ref{fig:keypoint_all} (A), the pink boxes illustrate a case where all four joints are selected once, which matches the video frame showing the ball at mid-air height with no distinctive gesture yet performed, thus requiring uniformly distributed attention. In Figure~\ref{fig:keypoint_fitnesaqa&scatter_plot}-A, we present the keypoint selection patterns from the FitnessAQA dataset. Here, the knee joints receive the highest selection frequency (shown in orange and green), which aligns perfectly with the fact that the performer is executing squats, where knee articulation is the most critical component of the motion. These qualitative observations collectively demonstrate that our MCTS-based reasoning module consistently identifies the essential keypoints at each timestep, providing faithful and interpretable visual groundings that lead to the model’s final prediction.

\section{Conclusion}
In this work, we presented LVDR, a novel framework for interpretable fine-grained skill assessment in sports and surgical videos. By formulating visual reasoning as a diffusion process in a latent reasoning space and integrating keypoint-guided Monte Carlo Tree Search (MCTS), LVDR not only delivers accurate performance predictions but also reveals the underlying reasoning trajectories that lead to these assessments. Extensive experiments across diverse datasets demonstrate that our approach achieves competitive quantitative results while providing interpretable insights into the spatiotemporal patterns critical for skill evaluation. Looking forward, we will extend LVDR to handle multi-person scenarios, integrating richer contextual information, and exploring real-time interpretable assessment for live applications in high-stakes domains such as surgical training and elite sports.

\section*{Acknowledgement}
This work was supported in part by the Virginia Commonwealth University start-up fund.


%
%
\bibliographystyle{splncs04}
\bibliography{main}
\end{document}